# Towards a Multiagent Decision Support System for crisis Management

Fahem Kebair

**ABSTRACT**

The cirsis management is a complex problem raised by the scientific community currently. Decision support systems are a suitable solution for such issues, they are indeed able to help emergency managers to prevent and to manage crisis in emergency situations. However, they should be enough flexible and adaptive in order to be reliable to solve complex problems that are plunged in dynamic and unpredictable environments.

The approach we propose in this paper addresses this challenge. We expose here a modelling of information for an emergency environment and an architecture of a multiagent decision support system that deals with these information in order to prevent the occur of a crisis or to manage it in emergency situations. We focus on the first level of the system mechanism which intends to perceive and to reflect the evolution of the current situation. The general approach and experimentations are provided here.

**INTRODUCTION**

Natural and man-made disasters are permanent hazards for human beings since they may have harmful consequences for them and their properties. In order to brace such events, people must be efficient in their evaluations, their decision-making and their actions. They must therefore change and perform their classical crisis management methods by using new means. This is already realized and accepted as a high priority task by many organizations, governments and companies in Europe and all over the world (Cutter et al. 2003).

In this context, Decision Support Systems (DSS) have proved their ability to resolve such kind of problems. Our research work addresses this challenge. It lies in building a DSS that must be able to help emergency managers to deal with crisit and to provide them emergency management

---

RCRSS functions in cycles of seconds. One second in the simulation represents one minute in real world plans for avoiding or reducing the consequences of these crisis. However, DSSs are well known to be customized for a specific purpose and can rarely be reused. Moreover, they only support circumstances which lie in the known and knowable spaces and do not support complex situations sufficiently (French & Niculae, 2005). Thus, the system may be used in different subjects of studies with minor changes. In other words, it operates in a generic manner and relies on specific knowledge that are defined by experts of the domain. Furthermore, the system may adapt its behaviour autonomously by altering its internal structure and changing its behaviour to better respond to the change of its environment. The MultiAgent Systems (MAS) technology is an appropriate solution to achieve these objectives. Intelligent agents (Wooldridge, 2002) are able to self-perform actions and to interact with other agents and their environment in order to carry out some objectives and to react to changes they perceive by adapting their behaviours.

The proposed system is made up of several agents organizations, of which kernel is operating on three levels. In this paper we focus on the first level in which a factual agents organisation has as role to perceive and to manage facts occurred in a dynamic environment. This step is fundamental in the final assessment of the situation. Indeed, the system creates its own representation of the environment state in order to extract the significant facts that may reveal the existence of risks. For this, it compares the current situation with previous known ones stored as scenarios. That way, the system may have a generic and adaptive mechanism and may learn during its process.

We tested the approach on several cases of studies in order to validate it. The work presented here is addressed to the RoboCupRescue Simulation System (RCRSS) (Kitano, 2000) (RoboCupRescue, 2010). We provide here a brief description of this application and we present and discuss the results we obtained.

**DECISION SUPPORT SYSTEM ARCHITECTURE**

DSSs are interactive, computer-based systems that aid users in judgement and choice activities. They provide data storage and retrieval but enhance the traditional information access and retrieval functions with support for model building and model-based reasoning. They support framing, modelling, and problem solving (Druzdzel and Flynn, 2000). More precisely, the purposes of a DSS are the following (Holsapple and Whinston, 1996):
- Supplementing the decision maker,
- Allowing better intelligence, design, or choice,
- Facilitating problem solving,
- Providing aid for non structured decisions,
- Managing knowledge.

---

RCRSS functions in cycles of seconds. One second in the simulation represents one minute in real world

**Role of the DSS in crisis management**

A crisis is a turning point or decisive change in a crucial situation. The uncertainty of the outcome is large; it can result in a disaster or pass almost unnoticed. There are many more and less precise definitions of crisis. They do depend on specific applications and situational circumstances. In our context, we are interested in natural and technological crisis. Crisis management (also known as emergency management or emergency response) is a dynamic process that begins well before the occur of the critical event and continues over its conclusion. The process involves a proactive, responsive and reflective component. Each stage of a crisis poses challenges for managers and decision makers and requires a different approach depending on the phase in question. This process is complex and exceeds widely the human abilities. Thus, DSSs may help to manage this process. Indeed, the DSS we present here must insure the following functionalities:
- Evaluation of the current situation, the system must detect/recognize an abnormal event;
- Evaluation/Prediction of the consequences, the system must assess the event by identifying the possible consequences;
- Intervention planning, the system must help the emergency responders in planning their interventions thanks to an actions plan (or procedures) that must be the most appropriate to the situation.

**DSS architecture**

Fig. 1 shows the over-all architecture of the DSS. The kernel is the main part of the system and has as role to manage all the decision-support process. The environment of the system includes essentially the actors and Distributed Information Systems (DIS), and feeds permanently the system with information describing the state of the current situation. In order to apprehend and to deal with these information, the system needs specific knowledge related to the domain as ontologies and proximity measures.

The system evaluates the possible consequences of the situation by comparing it with past situations. The latter represent the knowledge we hold about the treated problem and are stored in a Scenarios Base (SB). Thereby, an analogical reasoning is used based on the following postulate: if a given situation A seems like a situation B, then it is likely that the consequences of the situation A will be similar to those of B. Consequently, the risk appeared in B becomes a potential risk of A.

RCRSS functions in cycles of seconds. One second in the simulation represents one minute in real world

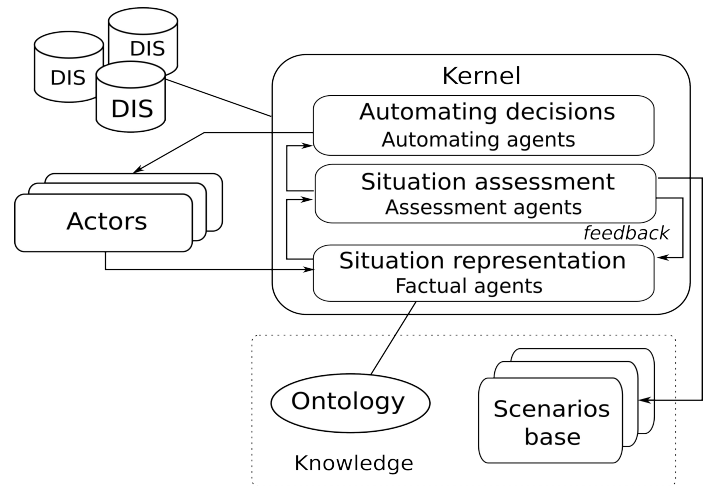

**Fig. 1:** Whole DSS Architecture

The kernel (see Fig. 1) is a MAS operating on three levels, it intends to detect significant organisations of agents in order to support finally the decision making. We intend, from such a structure, to equip the system with an adaptable and a partially generic architecture that may be easily adjusted to new cases of studies. Moreover, its suppleness allows the system to operate autonomously and to change its behaviour according to the evolution of the problem environment. The steps of the mechanism are detailed as follows:

- *Situation representation:* A fundamental step of the decision support process is to represent the current situation and its evolution over time. Indeed, the system perceives the facts that occur in the environment and creates its own representation of the situation thanks to a factual agents organisation. This approach has as purpose to let emerge subsets of agents.
- *Situation assessment:* A set of assessment agents are related to scenarios stored in a SB. These agents scrutinise permanently the factual agents organisation to find agents clusters enough close to their scenarios. This mechanism is similar to a Case-Based Reasoning (CBR) (Kolodner, 1993), except it is dynamic and incremental. According to the application, one or more most pertinent scenarios are selected to inform decision-makers about the state of the current situation and its probable evolution, or even to generate a warning in case of detecting a risk of crisis. The evaluation of the situation will be then reinjected in the perception level in order to confirm the position of the system about the current situation. This characteristic is inspired from the feedbacks of the natural systems. In this way, the system learns from its successes or its failures.
- *Automating decisions:* Outcomes generated by the assessment agents are captured by a set of automating agents and are transformed in decisions that may be used directly by the final users.

RCRSS functions in cycles of seconds. One second in the simulation represents one minute in real world

**RoboCupRescue case study**

We chose the RCRSS in order to apply the proposed approach. The RCRSS is an agent-based simulator which intends to reenact the rescue mission problem in real world. It reproduces an earthquake scenario which includes various kinds of incidents as the traffic after earthquake, buried civilians, road blockage, fire accidents, etc. A set of heterogeneous agents (RCR agents) coexist in the disaster space: rescue agents that are fire brigades, ambulance teams and police forces, and civilians agents. As in real case, RCR agents play the actors role here, they send their perceived information to the DSS in order to get a sequence of actions to perform. The DSS builds, based on these information, an overall knowledge which allows the evaluation of the whole situation. We focus, in this application, on the fires incidents and their related facts. We intend therefore to perceive and to represent both the fires propagation and the behaviour of the fire brigades.

A model of the RoboCupRescue disaster space and the properties of its components, and the RCR agents are detailed in (Takahashi, 2001). We use this model in order to extract knowledge and to formalise information.

**ENVIRONMENT DESIGN**

**Environment Modelling**

An important aspect of our approach is to model the observations issued from the environment. We defined a generic taxonomy in order to distinguish the different observed objects. This model represents a framework that helps to build an ontology whereof construction is often strenuous. Furthermore, it is used to specify the generic structure of the FSF in which will be written the observations reaching the system.

Drawing on diverse concrete cases, we propose a decomposition in six classes that may be qualified as abstract or generic (shown in Fig. 2). These classes belong to two families or-in other words-inherit two super classes. The first one comprises
concrete objects or exceptionable composite, it comprises three classes that describe
direct and concrete observations, concerning passive and active entities:
- Element objects are the components of the environment, e.g. buildings, roads, vegetation, . . .
- Person objects represent the actors of the environment;
- Group objects refer essentially to the used means and are generally a collection of various concrete objects, e.g. a car with its driver, a group of persons, . . .

The second family contains the virtual objects which are the different forms of activities; it gathers three other classes, deduced from indirect observations and are formalized based on the "memento" design pattern of Gamma (Gamma et al. 1995):

RCRSS functions in cycles of seconds. One second in the simulation represents one minute in real world

- Phenomenon objects are events that occur in the environment, e.g. fires, explosions, inundations, . . .
- Action objects are the activities that are performed by persons;
- Message objects represent the communication flow exchanged between persons.

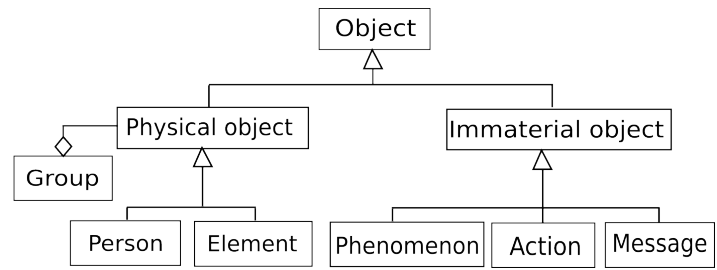

**Fig. 2:** Taxonomy of the perceived environment objects

**Ontology design**

In order to develop the DSS, we need to specify the knowledge related to the studied domain on which relies the decision-making. In literature, there is a distinction between data, information and knowledge. Data are defined as raw numbers and facts, information as processed data, interpreted into a meaningful context, and knowledge as being authenticated information made actionable (Vance, 1997). In our work, knowledge allows one to define the concepts, their characteristics and their relations, of the studied domain.

An approach to represent knowledge in a formal way is to build an ontology, which is an explicit specification of a conceptualization (Gruber, 1993). In our context, we use the taxonomy to build the specific ontology that represents the concepts and their relations in the crisis management domain. Moreover, we used the ontology to define similarities between the concepts. In fact, we defined semantic proximities between entities. More precisely, we defined an ontological graph where the nodes are the concepts and the arcs carry the proximities values that are beforehand fixed.

Distances here do not respect the property of the triangular inequality. They are normalized in an interval [-1,1]. When a similarity is 1 (i.e. exactly similar) the dissimilarity is -1.

**Factual semantic features**

The system receives and analyses permanently elementary information coming from the environment. These information are presented in an FSF shape. The noun given to this message content provides an explication to our approach: we stress observed and punctual elements that are the facts. A fact is a knowledge or information based on real occurrences It may be

---

RCRSS functions in cycles of seconds. One second in the simulation represents one minute in real world an event, an action, a phenomenon, etc.

Each FSF describes a fact and consequently a state change of an observed object issued from the taxonomy presented in (section). This may be modelled as a state-transition diagram. A transition represents an instantaneous transit from a state to another. It is triggered by an event (message), followed by the performance of one or several actions in the new state. The observation of this change is sent to the system in the shape of an FSF.

An FSF has a standard structure which is composed of <key,(qualifier, value)+>. The key is a unique identifier related to the observed object to which are associated some characteristics described by qualifiers and their respective values. Time and spatial values of the observation are also associated to an FSF. An example of an FSF is the following: <fire, intensity, strong, localization, building#12, time, 10:00 pm>. This fact describes a strong fire, located in building#12 and which is observed at 10:00 pm.

The homogeneity of the FSF structure has a great importance because it allows the system to deal with information in a generic way. This processing includes the update of information and their comparison with each other, which leads to make emerge the most significant facts of the observed situation. Comparing two FSFs consists in computing a proximity measure which is based on the semantic distance, discussed above. Thus, the more near to 1 the proximity measure between two FSFs is the more similar they are and vice versa. The semantic proximity Ps is computed using the ontology. Time and spatial proximities, noted respectively Pt and Pe, are computed using specific scales according to these formulas:

$$Pt = (4 \exp(-0.2\Delta t)) / (1 + \exp(-0.2\Delta t))^2$$
$$Pe = (4 \exp(-0.2\Delta e)) / (1 + \exp(-0.2\Delta e))^2$$

Where $\Delta t$ and $\Delta e$ are respectively the difference of time and the euclidean distance between the two observations.

The total proximity between two FSFs is the product of the three:
Proximity(FSF1, FSF2) = Ps x Pt x Pe

**REPRESENTATION MAS DESIGN: FACTUAL AGENTS**

**Structure**
Referring to the definition of an agent given by Wooldridge in (Wooldridge 1994), a factual agent is a reactive and a proactive agent. Each one carries an FSF and has as role to manage its evolution over time. We introduce the factual agents notion in the representation

---

RCRSS functions in cycles of seconds. One second in the simulation represents one minute in real world situation level to reflect the dynamic change of the situation on the one hand and to let emerge agents subsets on the other hand. These subsets may be representative of some situations that are close to some others encountered in the past. The final objective of the system is therefore to recognize and to evaluate these subsets.

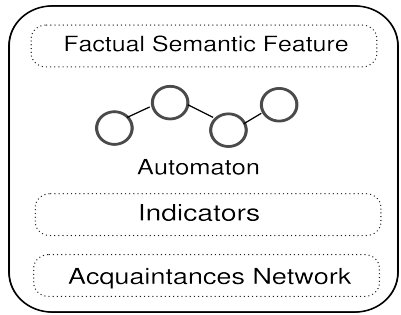

**Fig. 3:** Internal structure of a factual agent

A factual agent has specific indicators to reflect its dynamics. These indicators provide a synthetic view of the salient facts of the situation. They must reflect therefore as much as possible the perceived reality.

The behaviour of a factual agent is managed by an Augmented Transition Network (ATN) (Woods, 1970). The structure of the ATN is generic, however the conditions and the actions that are related respectively to the transitions and the states, are specific and depend on the FSF type of the factual agent.

**Interactions**
Factual agents are permanently in interaction and compare the FSFs that they carry with each other. A factual agent may have close agents (positive proximity between the FSFs) and opposite agents (negative proximity between the FSFs) and agents with which is neutral (proximity equals 0). It stores its close agents and opposite agents in an Acquaintances Network (AN) which is updated dynamically. We distinguish there kinds of messages, illustrated in the table below.

| FIPA Performative | Message |
|---|---|
| inform | FSFMessage |
| inform | AidMessage |
| inform | AgressionMessage |

— FSFMessage are messages that include an FSF and are sent by the agents, either when a new agent is created and wants to notify its creation to all the other agent, or when and agent updates its own FSF and wants

---

RCRSS functions in cycles of seconds. One second in the simulation represents one minute in real world to notify this change.

— AidMessage and AgressionMessage are messages that include numerical values, positve for the first and negative for the second, and are sent by the agents respectively to help a close agent or to agrees an apposite agent. Generally, these two actions are implemented in the ATN transitions, for example an agent may send both an AidMessage and an AgressionMessage when it transits from state 2 to state 3, or from state 3 to state 4, which confirms the imposing strength status of the agent.

**Behavioural activities**

The analysis of the factual agents groups is based on geometric criteria, insuring thus the independence of the treatment from the subject of study. Each factual agent exposes behavioural activities that are characterized thanks to numerical indicators. The definition of the latter and the way in which they are computed depend however on the treated application, since they must reflect as much as possible the described reality. The indicators form a behavioural vector that draws, by its variations, the dynamics of the agent during its live. This gives a meaning to the state of the agent inside its organization and consequently to the prominence of the semantic character that it carries.

**IMPLEMENTATION AND EXPERIMENTATION OF THE REPRESENTATION MAS**

**ATNs implementation**

To illustrate the evolution of both the state of a factual agent and its indicators, we discuss next two examples related to two factual agents related respectively to a fire phenomenon and a fire brigade agent.

Fig. 5 shows the two ATNs of these two factual agents. An ATN reflects the behaviour of the observed object represented by the factual agent, so a state change inside the ATN may have a meaning. However, the ATN does not describe exactly the state of this object in the real world.

Each factual agent has an automaton in four states. Both agents have a *Creation* state (state 1) in which the agent is created and enters in activity, and an *End* state (state 4), that means the agent dead. More precisely, a fire factual agent is dead when the fire is completely extinguished or when is burned, and a fire brigade factual agent is dead when the hit point of the fire brigade equals 0. Thus, the main states of these two factual agents are state 2 and state 3, in which they are active. A factual agent progress in this sens: 1 > 2 > 3 > 4 and regress in the opposite sens. The more the agent progress in its ATN, the more it acquires importance and a strength in its organisation.

---

RCRSS functions in cycles of seconds. One second in the simulation represents one minute in real world

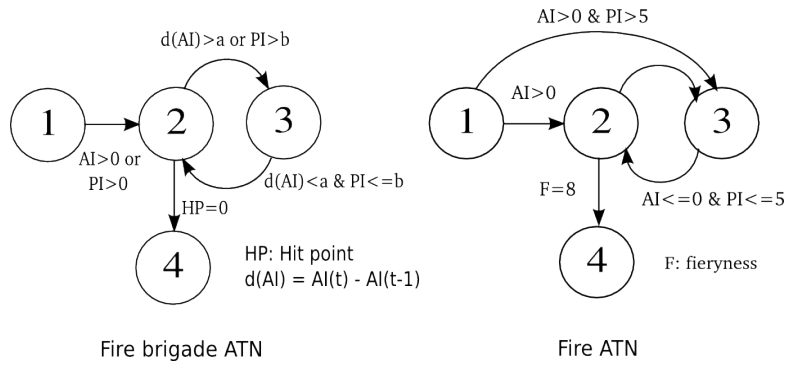

**Fig. 4:** ATNs of the the fire brigade and the fire factual agents

Both fire and fire brigade factual agents change state when their indicators values satisfy the transitions conditions. The latter are specific and depend on the type of the factual agent. They are defined as thresholds that may vary over time to allow the system to be more adaptive to the current situation.

**Indicators implementation**

Two kinds of factual agents have been developed: fire brigade factual agents and fire factual agents. The first ones describe facts related to the fire brigade agents of the RCRSS. The second ones reflect the fires evolution in the disaster space. Each agent has two indicators to reflect its dynamics. These indicators provide a synthetic view of the salient facts of the situation.

- *Action Indicator (AI):* it represents the position and the strength of a factual agent inside the representation MAS. For factual agents related to RCR agents, AI means the potential of an RCR agent and its efficiency in solving a problem. For factual agents managing phenomena, AI means the degree of damage and hazard that could represent this phenomenon.
- *Plausibility Indicator (PI):* for factual agents related to RCR agents, PI means the ability of an RCR agent to discover new problems in the disaster space. For phenomena factual agents, PI means the solving probability and the worsening impediment of a phenomenon.

As follows a way to compute AI and PI for a fire factual agent:

$AI = proximityMeasure(FSF1, FSF2)$

$PI = 10\, e^{-0.05Y}$
where $Y = [burningNeighbors + fieryness + lifeTime] - nbFireBrigades\}$;
lifeTime: time since the creation;
nbFireBrigades: number of fire brigades around the fire.

---

RCRSS functions in cycles of seconds. One second in the simulation represents one minute in real world

**Activities analysis**

In figure~\ref{fig:activites75}, the white chart illustrates the activities number of the representation MAS during a whole scenario. The activities include the states changes, the indicators values variations and the messages sent by the factual agents. The gray area represents the fire spreading, expressed by the number of the perceived fires over time. The system reacts in a moderate way at the beginning of the scenario, in which the fires are isolated. By dint of receiving more and more information, describing the fires propagation and the mobilization of the fire brigades, the factual agents react by intensifying their activities. The value and the oscillations of the activities number depend strongly on the behaviors of the factual agents. Indeed, the activities number grows when the fire brigades are fighting fires. Inversely, it drops when the fire brigades are potentially far from fires or are searching new ones.

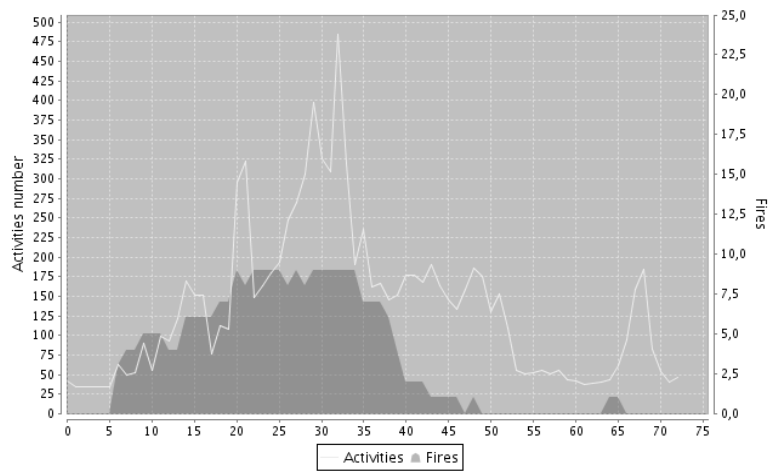

**Fig. 4:** Factual agents activities

At the end of the scenario, the system knows an evident bending result of the fires extinction. The factual agents become less pregnant since there are not important facts related to fires that come stimulating them. However, the system still in warning state in order to alert every notable change in the environment. We may notice this at the $63^{th}$ second of the simulation, when a fire reappears suddenly. The system reacts immediately to this fact and resumes its activities, then it becomes again stable after the fire were put out.

CONCLUSION

We proposed in this paper a part of an agent-based system that intends to

---

RCRSS functions in cycles of seconds. One second in the simulation represents one minute in real world help emergency managers to detect risks and to manage crisis situations. The main goal of our approach is to create a system that must be independent from the subject of study and that must be able to adapt autonomously its behaviour according to the environment change. We described here an original idea, using an agents organisation that allows the system to perceive occurred facts and to create its own representation of the situation. The final aim of the system is to recognize situations and to inform users about their potential consequences. We demonstrated the ability of the factual agents to react and to change their behaviours according to the sensed hazard.

Our current work concerns the creation of the factual agents clusters using the assessment agents and the way they will be stored and managed in the base. Therefore, a rigorous formalisation of clusters as well as distances measures to allow their comparison should be set up.

REFERECNES

RCRSS functions in cycles of seconds. One second in the simulation represents one minute in real world

RCRSS functions in cycles of seconds. One second in the simulation represents one minute in real world